\documentclass[10pt,journal]{IEEEtran}

\usepackage{booktabs}
\usepackage{multirow}
\usepackage{amsmath}
\usepackage{amssymb}
\usepackage{xcolor}
\usepackage{graphicx}
\usepackage{makecell}
\usepackage{cite}
\usepackage[breaklinks=true]{hyperref}

\title{SAGE: Saliency-Aware Generalist Encoder with\\Multi-level Feature Enhancement for Visual Recognition}
\author{Junyu Chen, Yucheng Huang, Le Yang, Qi Dang, Xin Kong, Haifeng Huo, and Qiang Zhao\thanks{Le Yang is the corresponding author. Email: yangle47@huawei.com}\\
Huawei Technologies Co., Ltd.}

\begin{document}
\maketitle

\begin{abstract}

Pretrained vision transformers (ViTs) have demonstrated strong transferability across visual recognition tasks. However, adapting a shared representation to diverse downstream tasks remains challenging, as different tasks demand varying levels of semantic abstraction and spatial precision. We present SAGE (Saliency-Aware Generalist Encoder), an encoder-side architecture that enhances pretrained ViT representations by integrating multi-depth features, register-token context, saliency-guided modulation, and aligned local views into a unified multi-scale feature pyramid. Evaluated with ViT-B and ViT-L backbones on six visual tasks across seven benchmarks, SAGE achieves 62.8 box AP, 55.3 mask AP, and 78.5 keypoint AP on COCO; 60.3 semantic segmentation mIoU and 44.9 instance segmentation AP on ADE20K; 94.76\% classification top-1 accuracy on TinyImageNet; and 81.10 oriented detection AP$_{50}$ on DIOR-R. Combined with a progressive two-stage training strategy, SAGE-L achieves the best performance among the selected reference methods in six of the seven benchmark settings, while remaining competitive on keypoint estimation against substantially larger models. Importantly, these tasks share the same SAGE encoder and multi-scale feature interface, while employing task-specific prediction heads. To the best of our knowledge, SAGE is the first publicly reported encoder to achieve highly competitive results on six visual tasks across seven benchmarks, demonstrating its generality and competitiveness.

\end{abstract}

\section{Introduction}
\label{sec:intro}
Pretrained vision transformers (ViTs) provide representations that transfer across image-level and spatial prediction tasks~\cite{dosovitskiy2021vit,caron2021dino,oquab2024dinov2}. Classification emphasizes image-level semantics, whereas detection, segmentation, and pose estimation additionally require spatially precise predictions. Adapting a pretrained ViT across these tasks therefore involves organizing its representations into features suited to different prediction heads.

Existing approaches establish effective interfaces for this adaptation. ViTDet constructs a feature pyramid from a single-scale representation~\cite{li2022vitdet}, DPT assembles intermediate transformer features at multiple resolutions~\cite{ranftl2021dpt}, and ViT-Adapter introduces spatial priors and repeated feature interactions~\cite{chen2023vitadapter}. Despite the impressive results achieved by the aforementioned works on vision tasks, ViT-based encoders still face the following challenges:

\begin{figure}[t]
\centering
\IfFileExists{compare.pdf}{\includegraphics[width=0.94\columnwidth]{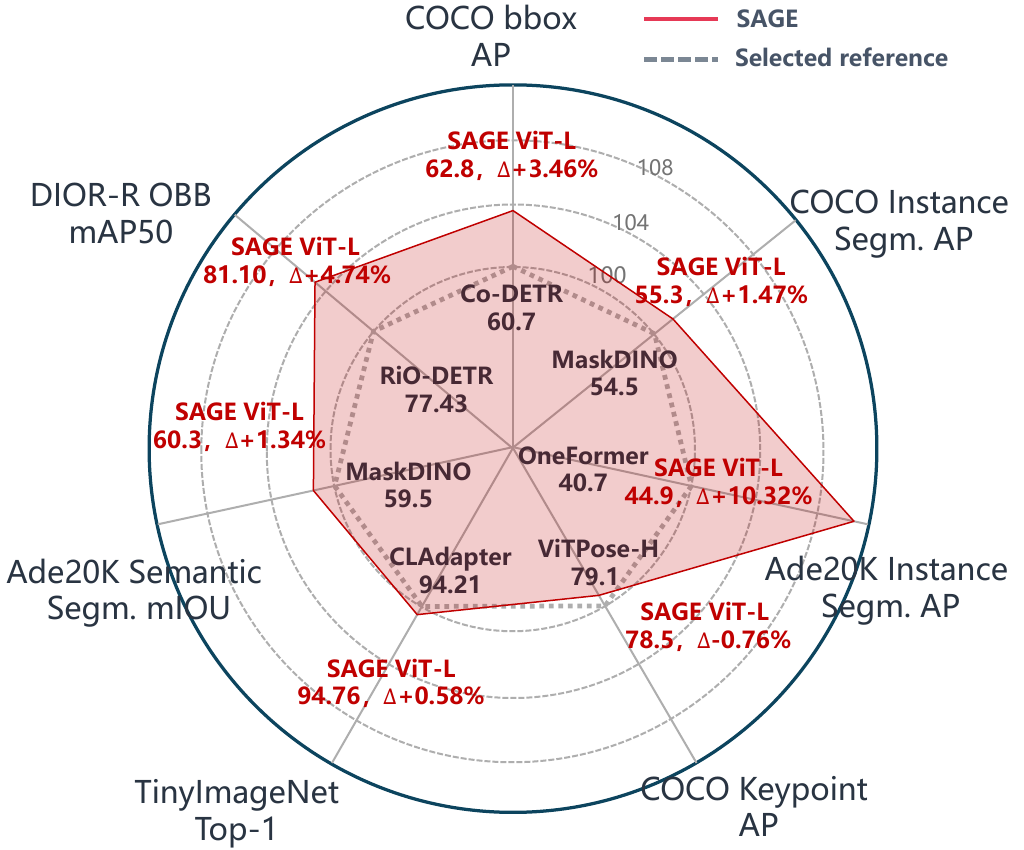}}{\fbox{\parbox[c][1.8in][c]{.92\columnwidth}{\centering\small Original figure location\\[4pt]\texttt{\detokenize{compare.pdf}}}}}
\caption{Cross-task performance of SAGE-L relative to the selected references in Table~\ref{tab:cross_task_main}. Each reference is normalized to 100\%; red labels give SAGE-L scores and relative percentage changes. Broader comparisons are reported in the supplement.}
\label{fig:fig_com}
\end{figure}

\textbf{Fine-Grained Spatial Representation and Adaptive Feature Enhancement.} A plain ViT operates on a coarse patch grid, which limits its ability to preserve fine-grained spatial structures required for precise localization and boundary prediction. Although approaches such as ViT-Adapter incorporate convolutional spatial priors to complement transformer representations, effectively integrating these heterogeneous features remains challenging. Moreover, while self-supervised ViTs encode rich global semantic information in the CLS token~\cite{caron2021dino}, such global representations alone do not provide explicit spatial guidance for adaptively enhancing individual feature locations.

\textbf{Comprehensive Utilization of Multi-Layer Features and Auxiliary Tokens.} ViTs encode complementary representations at different depths, while register tokens provide additional contextual information beyond patch tokens~\cite{darcet2024registers}. Existing adaptation methods typically aggregate selected intermediate features or employ predefined feature interactions~\cite{ranftl2021dpt,chen2023vitadapter}. However, they do not explicitly model scale-specific contributions from different transformer depths or exploit register-token representations for adaptive channel-wise modulation.

\textbf{Towards General-Purpose Visual Encoders.} Despite the remarkable performance of existing approaches across diverse vision tasks, achieving competitive results often relies on task-specific encoder–decoder architectures, with both components carefully designed and jointly optimized for individual tasks~\cite{carion2020detr,li2023maskdino,xu2022vitpose}. Such task-dependent designs tightly couple visual representation learning with downstream task requirements, limiting the flexibility and reusability of learned representations across different tasks. This raises an important question: how can we develop a general-purpose visual encoder that provides versatile representations for diverse vision tasks, without relying on task-specific architectural modifications?

To address these challenges, we propose SAGE, a general-purpose visual encoder designed to enhance visual representations and facilitate their transfer across diverse recognition tasks. Rather than relying on task-specific encoder–decoder co-design, SAGE aims to establish a unified representation learning framework that can effectively serve different downstream tasks. The central idea is to enhance the spatial expressiveness of pretrained ViT representations while comprehensively exploiting their hierarchical and contextual information. Our main contributions are summarized as follows:

\begin{itemize}
\item \textbf{A general-purpose visual encoder for diverse recognition tasks.} We propose SAGE, a unified encoder architecture that transforms pretrained ViT representations into a spatially expressive multi-scale feature pyramid, providing a shared representation interface for diverse visual recognition tasks. One encoder design covers six visual tasks across seven benchmarks, avoiding task-specific encoder architectures.
\item \textbf{A saliency-aware multi-level feature enhancement framework.} We introduce complementary mechanisms for spatial-prior-guided feature modulation, register-conditioned channel adaptation, multi-depth feature aggregation, and difference-aware local-view fusion, enabling more effective exploitation of semantic and fine-grained spatial information. The shared interface combines spatial detail and global context for downstream heads.
\item \textbf{A progressive two-stage training strategy.} We develop a quality-aware supervision schedule that progressively transitions from focal loss to Varifocal Loss, together with auxiliary feature modulation, to facilitate stable feature learning and improve downstream performance.
\end{itemize}

\begin{figure*}[!t]
\centering
\IfFileExists{SAGE_Figures_Editable.pdf}{\includegraphics[width=0.95\textwidth]{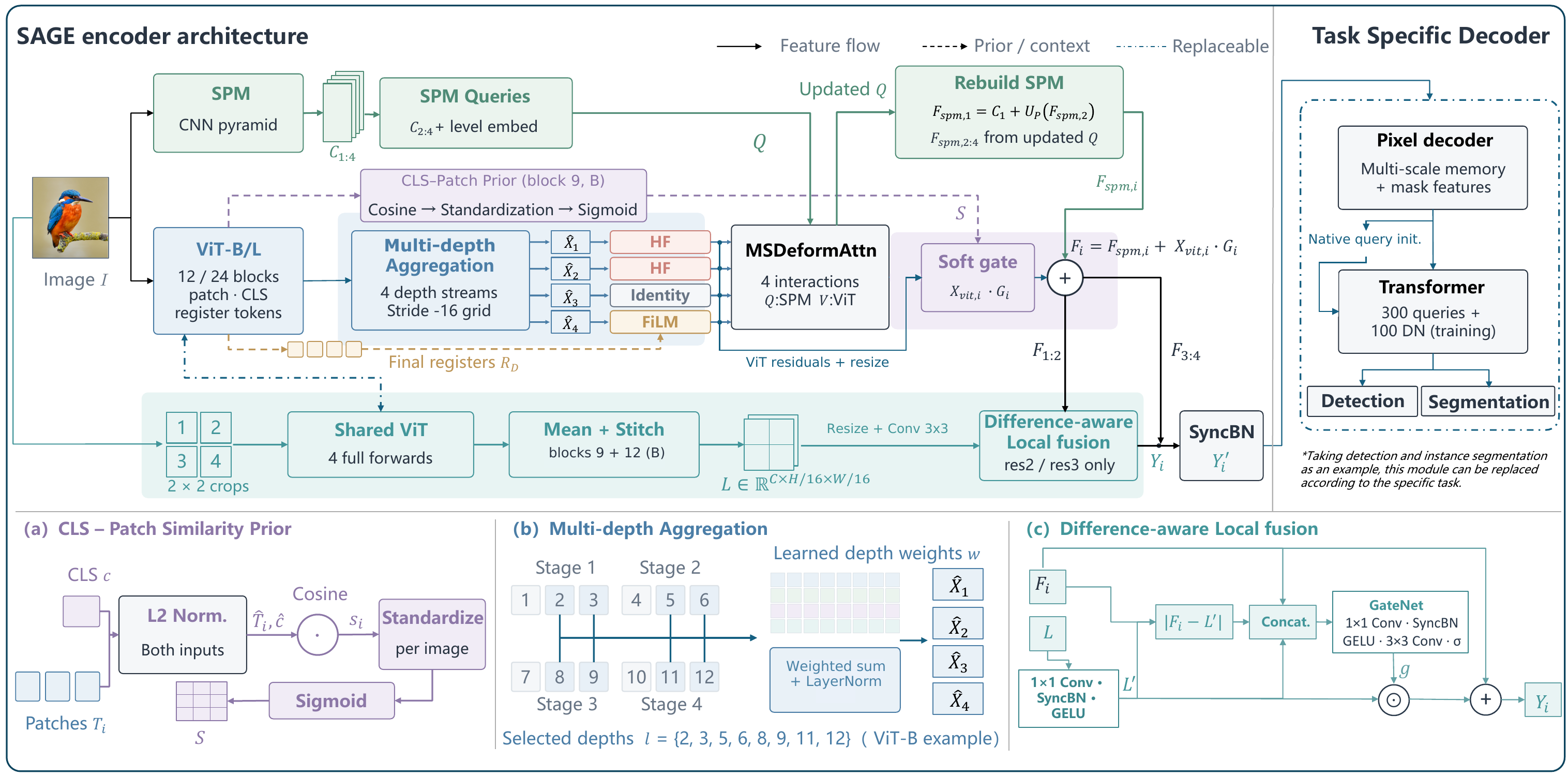}}{\fbox{\parbox[c][3.25in][c]{.96\textwidth}{\centering\small Original figure location\\[4pt]\texttt{\detokenize{SAGE_Figures_Editable.pdf}}}}}
\caption{Overview of SAGE. The Spatial-Prior Module(SPM) supplies multi-scale features; The ViT provides intermediate representations, CLS--patch correspondence, and register-token context; A local branch supplies features from four image views. Scale-specific aggregation and modulation prepare the ViT features for interaction with the spatial-prior path. The spatial-prior query($Q$) path is updated by attention, while the retained ViT features form a separate gated residual. Local-view fusion completes the pyramid consumed by the task head.}
\label{fig:fig1}
\end{figure*}

\section{Related Work}
\label{sec:related}
\subsection{Vision Transformer Representations and Adaptation}
\label{sec:detr_vit}
Self-supervised learning provides transferable ViT representations through increasingly large-scale pretraining~\cite{caron2021dino,oquab2024dinov2,simeoni2025dinov3}. Downstream adaptation determines how these representations are exposed to prediction heads. DPT reassembles intermediate tokens at multiple resolutions~\cite{ranftl2021dpt}; ViTDet constructs a simple feature pyramid from a plain ViT~\cite{li2022vitdet}; and ViT-Adapter introduces spatial priors and repeated interactions between transformer and spatial features~\cite{chen2023vitadapter}. These approaches establish multi-scale adaptation as an effective interface for spatial prediction. However, predefined feature interfaces do not learn scale-specific mixtures over a common set of transformer depths~\cite{li2022vitdet,ranftl2021dpt,chen2023vitadapter}.

DETR established end-to-end set prediction for object detection~\cite{carion2020detr}. Deformable DETR adds multi-scale deformable attention~\cite{zhu2021deformable}, DINO improves training through contrastive denoising~\cite{zhang2023dino}, and MaskDINO unifies detection and segmentation~\cite{li2023maskdino}. For pretrained ViTs, DEIMv2's spatial tuning adapter combines selected transformer layers with a lightweight CNN detail path and feature fusion~\cite{huang2025deimv2}. RF-DETR uses DINOv2 features, a multi-scale projector, and neural architecture search~\cite{robinson2026rfdetr}. However, these adaptation interfaces do not explicitly condition channel modulation on register-token context. Register tokens provide global information~\cite{darcet2024registers}, while FiLM provides a general feature-wise conditioning operation~\cite{perez2018film}. SAGE incorporates this contextual modulation into a shared encoder for diverse task-specific heads.

\subsection{{General-Purpose Encoders and \mbox{Training Strategies}}}
\label{sec:training_strategies}
General-purpose vision models seek reusable representations and prediction interfaces across tasks. Mask2Former introduces a common architecture for semantic, instance, and panoptic segmentation~\cite{cheng2022mask2former}, while OneFormer uses task-conditioned learning to support all three tasks within a single model~\cite{jain2023oneformer}. Their demonstrated task coverage remains confined to segmentation. SAGE pursues generality at the encoder level: spatial, contextual, and multi-depth information are integrated into a shared feature-pyramid design that can be paired with established task-specific heads. This design supports classification, detection, segmentation, and pose estimation, with task-specific objectives and training.

Focal loss addresses the imbalance between easy and hard examples~\cite{lin2017focal}. VarifocalNet introduces IoU-aware classification scores through Varifocal Loss (VFL)~\cite{zhang2021varifocalnet}, and DEIM proposes Matchability-Aware Loss to accommodate matches of varying quality~\cite{huang2025deim}. Positive supervision in VFL remains dependent on localization IoU~\cite{zhang2021varifocalnet}. SAGE progressively transitions from focal to varifocal supervision during detector training, complementing the shared encoder design.

\section{Method}
\label{sec:method}
\subsection{{Overall Architecture}}
\label{sec:overall}
SAGE combines three paths in a four-scale pyramid (Fig.~\ref{fig:fig1}). \textbf{SPM path:} the spatial-prior module~\cite{chen2023vitadapter} supplies CNN features $\mathbf C_{1:4}$ at strides 4, 8, 16, and 32. Tokens from strides 8--32 form the query sequence, which is updated by four successive attention interactions using ViT features as values. The highest-resolution feature is reconstructed as $\mathbf F_{\mathrm{spm},1}=\mathbf C_1+U_p(\mathbf F_{\mathrm{spm},2})$. \textbf{ViT path:} learned depth mixtures, Laplacian refinement, and register-conditioned FiLM prepare four token streams for these interactions. The same ViT streams are separately resized and spatially gated as residuals added to the updated SPM features. \textbf{Local path:} four image crops are processed by the shared ViT, spatially stitched, and fused into the two higher-resolution levels. SyncBN normalizes the resulting pyramid for the task head.

Section~\ref{sec:core1} describes the six components of the proposed feature adaptation framework, while Section~\ref{sec:training} introduces the corresponding training strategy. Task-specific heads and objectives are detailed in Section~\ref{sec:impl}.

\subsection{{Saliency-Aware Multi-Level Feature Enhancement}}
\label{sec:core1}
The adaptation module converts three sources---spatial-prior features, intermediate ViT representations, and local views---into the four-scale features consumed by the task head. The six components below specify this shared data flow.

\subsubsection{{DINO-style Saliency Computation}}
\label{sec:saliency}
Given intermediate patch tokens $\mathbf T\in\mathbb R^{N\times C}$ and a CLS token $\mathbf c\in\mathbb R^C$, we compute a spatial prior in three steps (Fig.~\ref{fig:fig1}(a)). First, we normalize the tokens and compute cosine similarity:
\begin{equation}
\begin{aligned}
\hat{\mathbf T}_i&=\frac{\mathbf T_i}{\|\mathbf T_i\|_2+\epsilon},\qquad
\hat{\mathbf c}=\frac{\mathbf c}{\|\mathbf c\|_2+\epsilon},\\
s_i&=\langle\hat{\mathbf T}_i,\hat{\mathbf c}\rangle.
\end{aligned}
\label{eq:saliency_cosine}
\end{equation}
Second, we standardize the similarities within each image:
\begin{align}
\mu&=\frac{1}{N}\sum_{i=1}^{N}s_i,\notag\displaybreak[3]\\
\sigma_{\mathrm{std}}&=\sqrt{\frac{1}{N}\sum_{i=1}^{N}(s_i-\mu)^2+\epsilon},\notag\displaybreak[3]\\
s'_i&=\frac{s_i-\mu}{\sigma_{\mathrm{std}}}.
\label{eq:saliency_standardization}
\end{align}
Third, sigmoid activation maps these values to a bounded prior:
\begin{equation}
S_i=\sigma(s'_i)=\frac{1}{1+e^{-s'_i}}.
\label{eq:spatial_prior}
\end{equation}
Here $\epsilon$ ensures numerical stability and $\sigma$ denotes sigmoid. For ViT-B, both tokens come from block 9 (one-based numbering). We reshape the $N=H_tW_t$ values to the token grid and resize the map to each pyramid scale. This annotation-free prior draws on spatial structure in self-supervised ViT features~\cite{caron2021dino}; constant and shuffled maps provide training-time controls.

\subsubsection{{Scale-Adaptive Residual Soft Gating}}
\label{sec:gating}
For each pyramid level $i$, a gate centered at one allows amplification and attenuation of the ViT residual:
\begin{equation}
\mathbf G_i=\operatorname{clamp}\!\left(
1+s_{\max}\alpha_i(2\mathbf S_i-1),\,1\pm s_{\max}\right),
\label{eq:gating}
\end{equation}
where clamp clips to $[1-s_{\max},1+s_{\max}]$, with $s_{\max}=0.20$. The strengths $\alpha_i=\sigma(\beta_i)$ are initialized by group to $\{0.20,0.50,0.80\}$: strides 4/8 share a group, while strides 16 and 32 have separate groups. Values above or below $S_i=0.5$ amplify or attenuate features without zeroing them. Only the ViT contribution is gated:
\begin{equation}
\mathbf F_i=\mathbf F_{\mathrm{spm},i}
+\mathbf X_{\mathrm{vit},i}\odot\mathbf G_i.
\label{eq:gated_feature}
\end{equation}
Leaving the SPM path ungated preserves its multi-scale spatial features.

\subsubsection{{Register-Token FiLM Modulation}}
\label{sec:film}
We explicitly use register-token context $\mathbf R\in\mathbb R^{K\times C}$~\cite{darcet2024registers} for feature-wise linear modulation (FiLM)~\cite{perez2018film}. Mean pooling and a two-layer MLP produce bounded channel-wise scale and shift parameters:
\begin{subequations}
\label{eq:film}
\begin{align}
\mathbf g_{\mathrm{ctx}}&=\operatorname{mean}(\mathbf R),\\
\mathbf h&=\operatorname{SiLU}(\mathbf W_1\mathbf g_{\mathrm{ctx}}+\mathbf b_1),\\
[\gamma_{\mathrm{raw}},\beta_{\mathrm{raw}}]&=\mathbf W_2\mathbf h+\mathbf b_2,\\
\gamma&=1+g_{\max}\tanh(\gamma_{\mathrm{raw}}),\\
\beta&=b_{\max}\tanh(\beta_{\mathrm{raw}}),\\
\mathbf y&=\mathbf x\odot\gamma+\beta.
\end{align}
\end{subequations}
With $g_{\max}=b_{\max}=0.10$, the scale lies in $[0.90,1.10]$ and the shift in $[-0.10,0.10]$. The transform approaches identity as the raw parameters approach zero. In ViT-B, four register tokens from block 12 condition the fourth token stream on its stride-16 grid, before its later alignment to output stride 32.

\subsubsection{{Laplacian High-Frequency Enhancer}}
\label{sec:highfreq}
Fine spatial structure is especially important at the shallow pyramid levels. We extract spatial variation in the ViT features using a fixed four-neighborhood Laplacian. Its kernel introduces no learned parameters:
\begin{equation}
\mathbf H=\operatorname{Lap}(\mathbf X)
=\mathbf X*\begin{bmatrix}0&1&0\\1&-4&1\\0&1&0\end{bmatrix}.
\label{eq:laplacian}
\end{equation}
By extracting local high-frequency responses and injecting them through a learnable residual, the module complements the semantic representations encoded by the ViT.
\begin{equation}
\mathbf Y=\mathbf X+\lambda\,s\,\operatorname{SiLU}(\mathbf H).
\label{eq:laplacian_residual}
\end{equation}
Here $s=0.20\,\sigma(s_{\mathrm{raw}})$ is initialized to 0.10; Section~\ref{sec:training} defines $\lambda$. This residual operates on the first two token streams at stride 16 ($64\times64$ for a $1024\times1024$ input), before their later alignment to output strides 4 and 8.

\begin{table*}[!t]
    \centering
    \caption{Comparison with state-of-the-art methods on COCO val2017. All reported results are single-modality (vision-only), and single-scale without test-time augmentation (TTA). Parameter counts are for the backbone. }
    \label{tab:sota}
    \tiny
    \setlength{\tabcolsep}{2pt}
    \renewcommand{\arraystretch}{0.9}
    \resizebox{\textwidth}{!}{
    \begin{tabular}{lccccccc}
    \toprule
    Method & Backbone & Params & Pretrain Data & bbox AP & box AP$_{50}$ & mask AP & mask AP$_{50}$ \\
    \midrule
    DINO-4scale (12 ep.)~\cite{zhang2023dino} & ResNet-50 & 25M & IN-1K & 49.0 & 66.6 & --- & --- \\
    MaskDINO~\cite{li2023maskdino} & ResNet-50 & 25M & IN-1K & 51.7 & \textemdash & 46.3 & 69.0 \\
    D-FINE-X~\cite{peng2025dfine} & HGNetv2-B5 & 30M & IN-1K + O365 & 59.3 & 76.8 & --- & --- \\
    RT-DETRv4-X~\cite{liao2025rtdetrv4} & HGNetv2 & \textemdash & IN-1K; VFM teacher & 57.0 & 74.6 & --- & --- \\
    DEIMv2-X~\cite{huang2025deimv2} & ViT-S+ & 29M & LVD-1689M & 57.8 & 75.4 & --- & --- \\
    DINO~\cite{zhang2023dino} & Swin-L & 197M & IN-22K & 58.5 & 78.0 & --- & --- \\
    ViTDet~\cite{li2022vitdet} & ViT-H & 632M & IN-1K (MAE) & 58.7 & \textemdash & 50.9 & \textemdash \\
    MaskDINO~\cite{li2023maskdino} & Swin-L & 197M & IN-22K & 58.3 & \textemdash & 52.1 & 76.5 \\
    MaskDINO (O365)~\cite{li2023maskdino} & Swin-L & 197M & IN-22K + O365 & --- & --- & \underline{54.5} & --- \\
    Co-DETR~\cite{zong2023codetr} & Swin-L & 197M & IN-22K & 60.7 & \textemdash & --- & --- \\
    RF-DETR-2XL (det.)~\cite{robinson2026rfdetr} & ViT-B & 86M & O365 & 60.1 & 78.5 & --- & --- \\
    RF-DETR-Seg-2XL~\cite{robinson2026rfdetr} & ViT-S & \textemdash & O365 & --- & --- & 49.9 & 73.1 \\
    \midrule
    \textbf{SAGE (ours)} & \textbf{ViT-B} & \textbf{86M} & \textbf{O365} & \underline{61.1} &   \underline{79.5} & 54.4 & \underline{78.5} \\
    \textbf{SAGE (ours)} & \textbf{ViT-L} & \textbf{304M} & \textbf{O365} & \textbf{62.8} & \textbf{81.3} &   \textbf{55.3} & \textbf{79.9} \\
    \bottomrule
    \end{tabular}
    }

    \end{table*}

\subsubsection{{Learned Multi-Layer Aggregation}}
\label{sec:dynamiclayer}
Different transformer depths provide distinct representations. Given selected layer outputs $\{\mathbf x_l\}_{l\in\mathcal I}$, we learn a separate mixture for each of the $S=4$ output scales (Fig.~\ref{fig:fig1}(b)):
\begin{equation}
\begin{aligned}
w_{s,l}&=\frac{\exp(W_{s,l}/\tau)}{\sum_{j\in\mathcal I}\exp(W_{s,j}/\tau)},\\
\hat{\mathbf x}_s&=\operatorname{LN}\!\left(\sum_{l\in\mathcal I}w_{s,l}\,\mathbf x_l\right).
\end{aligned}
\label{eq:aggregation}
\end{equation}
The mixing logits $W_{s,l}$ are learnable parameters shared across all input images, allowing the model to learn scale-specific preferences for transformer representations at different depths. We fix $\tau=2.0$ and initialize them with a distance-based depth prior:
\begin{equation}
W_{s,l}^{(\mathrm{init})}=-0.5\,|k(l)-c_s|,
\label{eq:aggregation_init}
\end{equation}
where $k(l)$ indexes the sampled positions. For ViT-B, $\mathcal I=\{2,3,5,6,8,9,11,12\}$ uses one-based block indices; $k(l)\in\{0,\ldots,7\}$ and $c_s\in\{1,3,5,7\}$. Patch and CLS features are aggregated separately and normalized after summation. All four streams retain the stride-16 token grid before subsequent processing and scale alignment.

\subsubsection{{Local Detail Branch and Difference-Aware Gated Fusion}}
\label{sec:localdetail}
To supply additional local-view features at the high-resolution pyramid levels (res2/res3), a $2\times2$ partition supplies four non-overlapping crops to the shared ViT without resizing the input crops. For a $1024\times1024$ image, each $512\times512$ crop yields a $32\times32$ token grid. In ViT-B, block-9 and block-12 features are averaged within each crop and stitched in image-coordinate order into a $64\times64$ map. 

The stitched map is resized to each high-resolution global feature $\mathbf F$ and projected by a $3\times3$ convolution, SyncBN, and GELU, giving $\mathbf L$. Inside the fusion block, a $1\times1$ projection gives
\begin{equation}
\mathbf L'=\mathcal P_{1\times1}(\mathbf L),\qquad
\mathbf D=|\mathbf F-\mathbf L'|,
\label{eq:local_projection}
\end{equation}
where $\mathcal P_{1\times1}$ includes SyncBN and GELU. The fusion gate combines both features and their discrepancy:
\begin{equation}
\begin{aligned}
\mathbf g&=\sigma\!\left(\operatorname{Conv}_{3\times3}
\bigl(\mathcal P_{1\times1}([\mathbf F,\mathbf L',\mathbf D])\bigr)\right),\\
\mathbf Y&=\mathbf F+\mathbf g\odot\mathbf L'.
\end{aligned}
\label{eq:local_fusion}
\end{equation}
The two $\mathcal P_{1\times1}$ projections have separate parameters. Fusion operates at output strides 4 and 8 (Fig.~\ref{fig:fig1}(c)); this local path and multi-layer aggregation remain active at inference.

\subsection{{Training Strategy}}
\label{sec:training}
\subsubsection{{Two-Stage Quality-Aware Supervision}}
\label{sec:core3}\label{sec:vfl}
VFL uses localization IoU as the classification target and as a positive-example weight~\cite{zhang2021varifocalnet}. Early localization quality can therefore affect positive supervision. We begin detector training with sigmoid focal loss~\cite{lin2017focal} ($\alpha=0.25$, $\gamma=2.0$), then progressively introduce VFL. For transition onset $t_0$ and length $R>0$, the mixing coefficient is
\begin{equation}
\rho(t)=\begin{cases}
0,&t<t_0,\\
(t-t_0)/R,&t_0\le t<t_0+R,\\
1,&t\ge t_0+R.
\end{cases}
\label{eq:vfl_schedule}
\end{equation}
The classification objective becomes
\begin{equation}
\mathcal L_{\mathrm{cls}}=(1-\rho(t))\mathcal L_{\mathrm{FL}}
+\rho(t)\mathcal L_{\mathrm{VFL}}.
\label{eq:vfl_objective}
\end{equation}
For a matched prediction, the VFL target is $\mathbf y_{\mathrm{onehot}}\operatorname{sg}(\mathrm{IoU})$, where $\operatorname{sg}$ stops gradients through the quality target. The VFL-from-start control uses $\rho\equiv1$. The pretraining study uses focal supervision for 10K steps, followed by a 1K linear transition and VFL through 20K. Section~\ref{sec:vfl_ablation2} evaluates the observed training trajectories.

\paragraph{Auxiliary modulation.}
\label{sec:core2}
We also propose an auxiliary modulation method for the training stage, where a training coefficient $\lambda(t)$ scales the auxiliary residuals: $\mathbf G_i^{(\lambda)}=1+\lambda(\mathbf G_i-1)$, $\gamma^{(\lambda)}=1+\lambda(\gamma-1)$, and $\beta^{(\lambda)}=\lambda\beta$. Equation~\ref{eq:laplacian_residual} already includes $\lambda$. This formulation controls the strength of auxiliary feature enhancement throughout training through the modulation function $\lambda(t)$. The reference schedule, independent aggregation-weight freezing, and bypass controls are documented in the supplement.

\subsubsection{{Task Objectives}}
\label{sec:loss}\label{sec:queryinit}\label{sec:semloss}
Detection and instance segmentation combine classification, localization, mask, and denoising objectives~\cite{li2023maskdino}:
\begin{equation}
\begin{aligned}
\mathcal L={}&\mathcal L_{\mathrm{cls}}+\mathcal L_{\mathrm{box}}
+\mathcal L_{\mathrm{mask}}+\mathcal L_{\mathrm{dice}}\\
&+\mathcal L_{\mathrm{dn}}+\lambda_{\mathrm{sem}}\mathcal L_{\mathrm{sem}}.
\end{aligned}
\label{eq:objective}
\end{equation}
Task-specific weights are absorbed into each term. Hungarian matching uses the weighted classification, mask, Dice, box, and generalized-IoU costs:
\begin{equation}
\begin{aligned}
\mathcal C={}&\lambda_{\mathrm{cls}}\mathcal C_{\mathrm{cls}}
+\lambda_{\mathrm{mask}}\mathcal C_{\mathrm{mask}}
+\lambda_{\mathrm{dice}}\mathcal C_{\mathrm{dice}}\\
&+\lambda_{\mathrm{box}}\mathcal C_{\mathrm{box}}
+\lambda_{\mathrm{giou}}\mathcal C_{\mathrm{giou}}.
\end{aligned}
\label{eq:matching}
\end{equation}
All controlled subset experiments retain native MaskDINO query initialization and set $\lambda_{\mathrm{sem}}=0$. The full-data cumulative study includes the semantic-consistency objective described in the supplement. Other tasks use their corresponding heads and objectives.

\begin{table*}[!t]
\begin{minipage}[t]{0.48\textwidth}

\centering
\caption{Module ablations on the fixed 20K-image COCO subset (ViT-B, 17K updates). $\Delta$ is Full minus the corresponding configuration.}
\label{tab:subset_ablation}
\small
\begin{tabular*}{\columnwidth}{@{\extracolsep{\fill}}llrr@{}}
\toprule
ID & Configuration & Box AP & $\Delta$\\
\midrule
E1 & Full {SAGE} & {56.46} & {--}\\
E2 & {w/o Local details} & {55.84} & {0.62}\\
E3 & w/o Saliency module & 56.01 & 0.45\\
E4 & w/o Gate + FiLM + Lap. & 56.32 & 0.14\\
E5 & {w/o FiLM} & {56.25} & {0.21}\\
E6 & {w/o Laplacian} & {56.42} & {0.04}\\
E7 & Baseline & 55.35 & 1.11\\
\bottomrule
\end{tabular*}

\end{minipage}\hfill
\begin{minipage}[t]{0.48\textwidth}

\centering
\caption{Cumulative ablation on COCO using ViT-B with 100K training iterations. $\Delta$ denotes the AP gain over the preceding configuration. Training strategy means 2-stage VFL + Auxiliary modulation.}
\label{tab:ablation}
\small
\begin{tabular*}{\columnwidth}{@{\extracolsep{\fill}}lcc@{}}
\toprule
Configuration & Box AP & $\Delta$ \\
\midrule
Baseline (w/ gate + FiLM + Lap.) & 57.70 & --- \\
+ Saliency module & 58.90 & +1.20 \\
+ Local details & 60.91 & +2.01 \\
+ Training strategy & \textbf{61.10} & +0.19 \\
\bottomrule
\end{tabular*}

\end{minipage}
\end{table*}

\section{{Experiments}}
\label{sec:exp}

\subsection{{Datasets and Evaluation Metrics}}
\label{sec:datasets}
We evaluate on four datasets: \textbf{COCO}~\cite{lin2014coco} (object detection, instance segmentation, and keypoints), \textbf{ADE20K}~\cite{zhou2017ade20k} (semantic and instance segmentation), \textbf{TinyImageNet}~\cite{le2015tinyimagenet} (classification), and \textbf{DIOR-R}~\cite{cheng2022aopg} (rotated detection). SAGE uses single-scale inference without test-time augmentation. Published reference configurations retain their reported training and evaluation settings, detailed in the supplement.

\subsection{{Implementation Details}}
\label{sec:impl}
\textbf{Encoder and task heads.} The SPM produces features at strides 4, 8, 16, and 32; SAGE is designed to improve ViT feature representations at the backbone level rather than through task-specific modules. To evaluate this design, we use existing or lightweight task heads where possible.  MaskDINO~\cite{li2023maskdino} decodes the shared encoder features for object detection and segmentation. Classification uses a two-layer MLP, and keypoint estimation aligns the four feature levels to stride 4 before a $1\times1$ prediction convolution. Oriented detection uses an adapted RiO-DETR~\cite{hu2026riodetr} head.

\textbf{Training protocols.} Full-data COCO training uses 100K iterations, batch sizes of 16--32, AdamW~\cite{loshchilov2019adamw} with learning rate $10^{-4}$ and weight decay 0.05, and WarmupPolyLR on Huawei Ascend 910B2 NPUs.

\begin{figure}
\centering
\includegraphics[width=0.94\columnwidth]{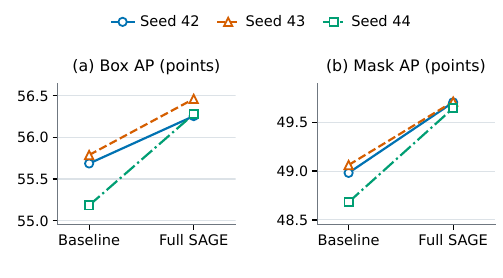}
\caption{Paired-seed results on the fixed 20K-image COCO subset after 17K updates. Each line connects Baseline and Full SAGE with the same seed. All three paired comparisons improve in box and mask AP.}
\label{fig:multiseed_paired}
\end{figure}

\begin{figure*}
\centering
\includegraphics[width=0.72\textwidth]{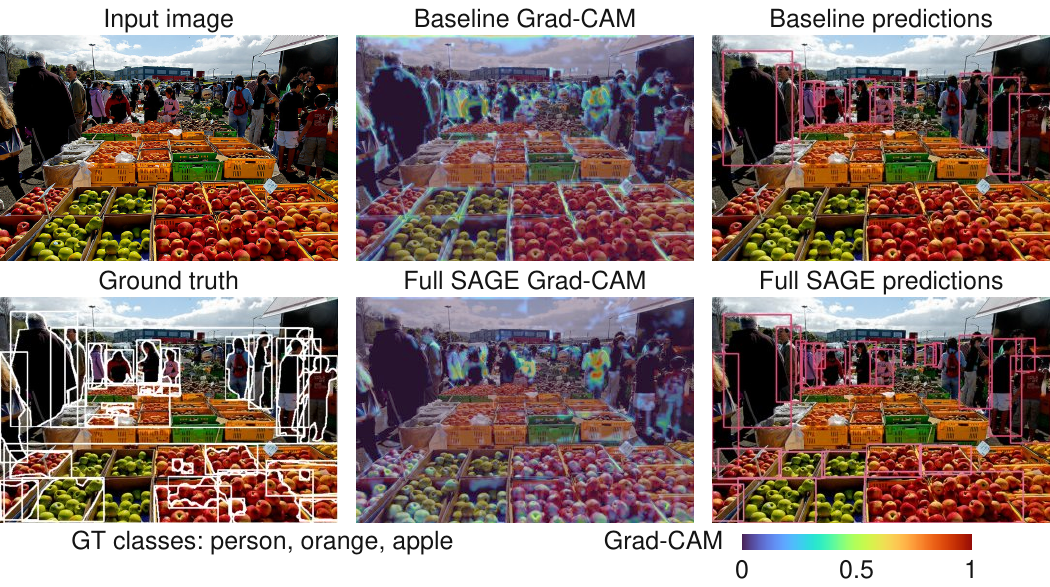}
\caption{Qualitative comparison on COCO val2017: input image, ground truth, Baseline/SAGE encoder Grad-CAM maps, and their predictions. Colors indicate normalized Grad-CAM values.}
\label{fig:encoder_qualitative}
\end{figure*}

\begin{figure*}
\centering
\includegraphics[width=0.65\textwidth]{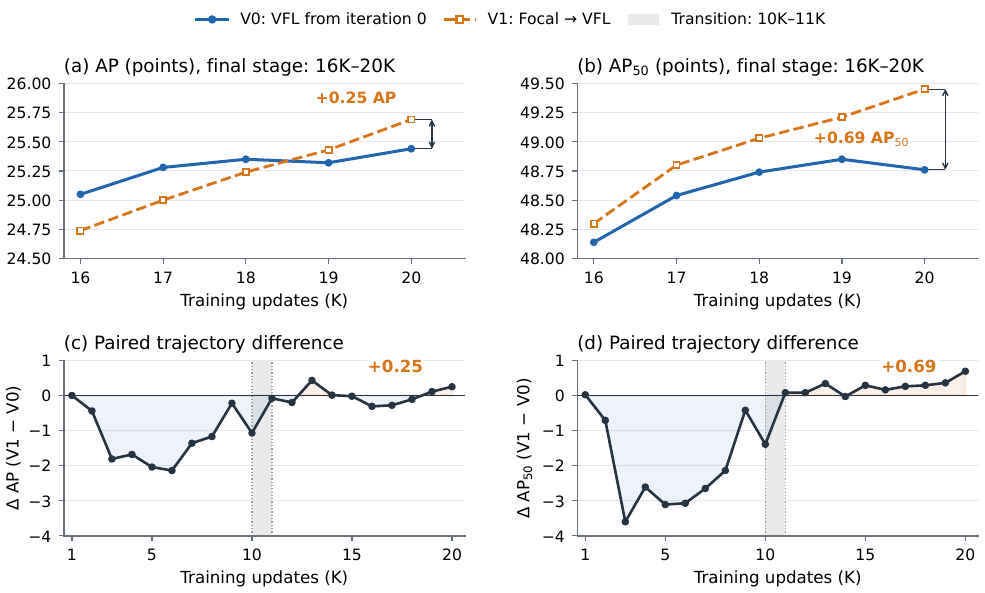}
\caption{Comparison of two VFL training schedules. V0 uses VFL from the first update. V1 uses focal loss for 10K updates, then transitions linearly to VFL during 10K--11K. (a, b) AP and AP$_{50}$ over 16K--20K; (c, d) V1$-$V0 differences over all 20K updates. Both runs learn successfully; V1 has higher final AP and AP$_{50}$.}
\label{fig:vfl_startup}
\end{figure*}

\subsection{{COCO Object Detection and Instance Segmentation}}
\label{sec:coco}
\subsubsection{{Comparison with SOTA Methods}}
\label{sec:sota}
Table~\ref{tab:sota} compares SAGE with state-of-the-art detectors and instance segmentation models. SAGE-L obtains 62.8 box AP and 55.3 mask AP on COCO val2017 with single-scale inference and no TTA. 

\begin{itemize}
\item \textbf{SAGE ViT-B vs. RF-DETR-2XL}: Both methods employ ViT-B-scale backbones, although their pretraining configurations differ. {SAGE}-B improves box AP over RF-DETR-2XL by 1.0 point (61.1 vs.\ 60.1) and mask AP over RF-DETR-Seg-2XL by 4.5 points (54.4 vs.\ 49.9). 

\item \textbf{SAGE ViT-B vs. Co-DETR}: SAGE-B surpasses Co-DETR with Swin-L by 0.4 box AP (61.1 vs.\ 60.7), using an 86M-parameter backbone compared with 197M for Swin-L. This result highlights the accuracy achieved by the compact SAGE-B encoder.
\end{itemize}

\begin{table*}[!t]
\centering
\caption{Cross-task comparison with representative image-only-trained reference configurations. SAGE uses single-scale inference without TTA; reference settings, larger image-only models, and image--text-pretrained comparisons are detailed in the supplement. Bold/underline mark the highest/second-highest displayed scores.}
\label{tab:cross_task_main}
\setlength{\tabcolsep}{2.5pt}
\renewcommand{\arraystretch}{0.86}
\resizebox{\textwidth}{!}{%
\begin{tabular}{clccccc|cc}
\toprule
\shortstack{Task /\\Dataset} & Method & Backbone & \shortstack{Backbone\\Params.} & \shortstack{Pretraining /\\Extra data} & Metric & Ref. & \shortstack{SAGE\\B} & \shortstack{SAGE\\L}\\
\midrule
\multirow{3}{*}{\shortstack{Detection\\(COCO)}} & D-FINE-X~\cite{peng2025dfine} & HGNetv2-B5 & 30M & O365 & \multirow{3}{*}{Box AP} & 59.3 & \multirow{3}{*}{\underline{61.1}} & \multirow{3}{*}{\textbf{62.8}} \\
 & RF-DETR-2XL~\cite{robinson2026rfdetr} & ViT-B & 86M & DINOv2 + O365 &  & 60.1 &  &  \\
 & Co-DETR~\cite{zong2023codetr} & Swin-L & 197M & IN-22K &  & 60.7 &  &  \\
\midrule
\multirow{3}{*}{\shortstack{Instance seg.\\(COCO)}} & RF-DETR-Seg-2XL~\cite{robinson2026rfdetr} & DINOv2-S & --- & DINOv2 + O365 & \multirow{3}{*}{Mask AP} & 49.9 & \multirow{3}{*}{54.4} & \multirow{3}{*}{\textbf{55.3}} \\
 & ViTDet~\cite{li2022vitdet} & ViT-H & 632M & IN-1K (MAE) &  & 50.9 &  &  \\
 & MaskDINO~\cite{li2023maskdino} & Swin-L & 197M & IN-22K + O365 &  & \underline{54.5} &  &  \\
\midrule
\multirow{3}{*}{\shortstack{Instance seg.\\(ADE20K)}} & Mask2Former + DiNAT~\cite{jain2023oneformer} & DiNAT-L & 200M & IN-22K & \multirow{3}{*}{Mask AP} & 35.4 & \multirow{3}{*}{\underline{43.3}} & \multirow{3}{*}{\textbf{44.9}} \\
 & OneFormer~\cite{jain2023oneformer} & DiNAT-L & 200M & IN-22K &  & 37.1 &  &  \\
 & OneFormer + ViT-P~\cite{jain2023oneformer,shahabodini2026vitp} & DiNAT-L + ViT-P & 300M & IN-22K + COCO &  & 40.7 &  &  \\
\midrule
\multirow{3}{*}{\shortstack{Semantic seg.\\(ADE20K)}} & Mask2Former~\cite{cheng2022mask2former} & Swin-L & 197M & IN-22K & \multirow{3}{*}{mIoU (\%)} & 56.1 & \multirow{3}{*}{58.7} & \multirow{3}{*}{\textbf{60.3}} \\
 & ViT-Adapter + UPerNet~\cite{chen2023vitadapter} & ViT-L & 304M & IN-22K (BEiT) &  & 58.0 &  &  \\
 & MaskDINO~\cite{li2023maskdino} & Swin-L & 197M & IN-22K + O365 &  & \underline{59.5} &  &  \\
\midrule
\multirow{3}{*}{\shortstack{Rotated det.\\(DIOR-R)}} & YOLO26x-OBB~\cite{wang2026yolo26,hu2026riodetr} & YOLO26x & 56M & --- & \multirow{3}{*}{AP$_{50}$} & 76.48 & \multirow{3}{*}{\underline{78.44}} & \multirow{3}{*}{\textbf{81.10}} \\
 & RT-DETRv2-OBB-X~\cite{lv2024rtdetrv2} & HGNetV2-B5 & 33M & IN-1K &  & 76.52 &  &  \\
 & RiO-DETR-X~\cite{hu2026riodetr} & HGNetV2-B5 & 33M & IN-1K &  & 77.43 &  &  \\
\midrule
\multirow{3}{*}{\shortstack{Classification\\(TinyImageNet)}} & DeiT-B/16-D~\cite{touvron2021deit,huynh2022vit} & DeiT-B/16-D & 87M & IN-1K & \multirow{3}{*}{Top-1 (\%)} & 87.29 & \multirow{3}{*}{91.46} & \multirow{3}{*}{\textbf{94.76}} \\
 & Swin-L/4~\cite{liu2021swin,huynh2022vit} & Swin-L/4 & 196M & IN-22K &  & 91.35 &  &  \\
 & CLAdapter~\cite{li2025cladapter} & ViT-L & 304M & IN-21K &  & \underline{94.21} &  &  \\
\midrule
\multirow{4}{*}{\shortstack{Keypoints\\(COCO)}} & RTMPose-L~\cite{jiang2023rtmpose} & CSPNeXt-L & 26M & IN-1K & \multirow{4}{*}{AP} & 75.8 & \multirow{4}{*}{76.0} & \multirow{4}{*}{\underline{78.5}} \\
 & ViTPose-B~\cite{xu2022vitpose} & ViT-B & 86M & IN-1K &  & 75.8 &  &  \\
 & ViTPose-L~\cite{xu2022vitpose} & ViT-L & 306M & IN-1K &  & 78.3 &  &  \\
 & ViTPose-H~\cite{xu2022vitpose} & ViT-H & 631M & IN-1K &  & \textbf{79.1} &  &  \\
\bottomrule
\end{tabular}}
\par\smallskip
\begin{minipage}{\textwidth}\footnotesize
SAGE-B/L use ViT-B/L backbones (86M/304M parameters) with Objects365~\cite{shao2019objects365} initialization. Parameter counts exclude task heads. 
\end{minipage}
\end{table*}

\subsubsection{{Fine-Grained Encoder Ablations}}
\label{sec:ablation}

To efficiently evaluate the contribution of each component, we conduct ablation experiments on a 20K subset of COCO, with each variant trained for 17K iterations under identical settings.  

As shown in Table ~\ref{tab:subset_ablation}, the full SAGE achieves the best performance of 56.46 AP. Removing Local Detail Enhancement causes the largest performance drop of 0.62 AP, highlighting the importance of preserving fine-grained local information. Similarly, replacing the saliency prior by its spatial mean lowers AP by 0.45, demonstrating the effectiveness of saliency-aware feature enhancement. Removing FiLM results in a 0.21 AP drop, supporting its role in adaptive feature modulation. Although removing Laplacian Enhancement leads to a relatively small decrease of 0.04 AP, we retain this lightweight operation to emphasize local spatial variations. Furthermore, jointly removing the Gating Mechanism results in a 0.14 AP drop.

Figure~\ref{fig:multiseed_paired} shows the within-seed comparisons. Across three paired random seeds, the baseline achieves $55.55\pm0.32$ box AP and $48.91\pm0.20$ mask AP, whereas full SAGE achieves $56.33\pm0.11$ and $49.69\pm0.04$, respectively. Notably, SAGE consistently outperforms the baseline in both metrics across all three seeds, yielding mean paired improvements of $0.78$ box AP and $0.78$ mask AP.

\par\smallskip\begingroup\clubpenalty=10000\noindent\textbf{Qualitative comparison.}\quad Figure~\ref{fig:encoder_qualitative} shows more concentrated Grad-CAM~\cite{selvaraju2017gradcam} responses on fruit instances for Full SAGE, whereas baseline responses are dispersed and emphasize crate boundaries. Full SAGE also detects fruit instances missed by the baseline, qualitatively supporting the effectiveness of our encoder enhancements in cluttered scenes.\par\endgroup

\subsubsection{{Cumulative Ablations}}
\label{sec:module_contributions}

The saliency and local-detail components show the largest individual gains in Table~\ref{tab:subset_ablation}. To further investigate the cumulative benefits of integrating different components, we conduct progressive experiments under a unified training setup, with the results reported in Table \ref{tab:ablation}. Starting from the baseline equipped with the Gating Mechanism, FiLM, and Laplacian Enhancement, 
incorporating saliency modulation improves box AP by 1.20 points. Integrating the local detail branch brings an additional improvement of 2.01 AP. Finally, introducing the training strategy, comprising two-stage VFL and auxiliary modulation, further yields a gain of 0.19 AP.

\subsubsection{{Training Dynamics of Two-stage VFL}}
\label{sec:vfl_ablation2}

To investigate the effect of VFL training schedules, we compare two strategies during detector pre-training on Objects365: applying VFL from the first iteration (V0) and adopting a two-stage schedule (V1), which employs Focal Loss (FL) during the initial 10K iterations, followed by a linear transition to VFL over 10K--11K iterations. Both strategies are evaluated every 1K iterations on the validation set over 20K training iterations. As shown in Fig. \ref{fig:vfl_startup}, V0 exhibits faster early learning during 2K--10K iterations. However, following the transition to VFL, V1 gradually closes the performance gap and eventually surpasses V0. At 20K iterations, V1 achieves 25.69 AP and 49.45 AP$_{50}$, compared with 25.44 AP and 48.76 AP$_{50}$ for V0, yielding gains of 0.25 AP and 0.69 AP$_{50}$, respectively. Detailed experimental settings are provided in the supplement.

\subsection{{Cross-task Generalization and Versatility}}
\label{sec:othertasks}
Table~\ref{tab:cross_task_main} evaluates the shared encoder architecture and corresponding task heads across seven benchmark settings spanning six visual task categories: object detection, instance segmentation, semantic segmentation, oriented object detection, image classification, and keypoint estimation. In particular, SAGE-L achieves the best results among the displayed references in six of the seven benchmark settings, highlighting its broad applicability and strong transferability. Consistently, SAGE-B/L reach 43.3/44.9 mask AP on ADE20K, and SAGE-B/L obtain 78.44/81.10 AP$_{50}$ on DIOR-R. For keypoints, SAGE-L reaches 78.5 AP, compared with 78.3 for ViTPose-L and 79.1 for ViTPose-H. SAGE-L achieves comparable performance with approximately half the backbone parameters (304M vs. 631M), demonstrating favorable parameter efficiency. SAGE-L achieves 60.3 mIoU on ADE20K semantic segmentation. Classification reaches 94.76\% top-1 accuracy, 0.55 points above the displayed CLAdapter result. These results demonstrate broad task applicability and strong accuracy at both backbone scales.

The cross-task comparisons evaluate complete systems; their pretraining, evaluation settings, and oriented-detection head configurations are detailed in the supplement. The supplement includes larger and image--text-pretrained reference configurations to contextualize the comparison range.

\section{{Conclusion}}
\label{sec:conclusion}

In this paper, we presented SAGE, a saliency-aware generalist encoder designed to enhance pretrained ViT representations for diverse visual recognition tasks. SAGE introduces a unified multi-level feature enhancement framework that integrates saliency-guided spatial modulation, register-conditioned channel adaptation, multi-depth feature aggregation, and local-view fusion. 

Beyond architectural improvements, we further investigated the interaction between feature enhancement and optimization, introducing a progressive two-stage VFL training strategy with auxiliary feature modulation. 

We evaluated SAGE across seven benchmark settings covering six visual recognition task categories. Experimental results show that SAGE-L achieves the best performance among the selected reference methods in six of the seven settings, while remaining competitive with substantially larger models on keypoint estimation. Both ViT-B and ViT-L variants demonstrate strong performance, highlighting the scalability and transferability of the proposed encoder. These findings motivate further development of general-purpose visual encoders.

\bibliographystyle{IEEEtran}
\bibliography{main}

\clearpage
\section*{Supplementary Material}
This supplement provides implementation details and additional experimental results. Section~\ref{supp:implementation} specifies the encoder, task interfaces, and experimental protocols. Section~\ref{supp:encoder_evidence} presents additional encoder ablations and paired-seed results. Sections~\ref{supp:training} reports training-strategy analyses. Section~\ref{supp:cross_task} extends the cross-task comparisons. All results for SAGE were obtained using single-scale inference without TTA, while the compared methods retained their respective training protocols.

\section{Implementation Details and Experimental Protocols}
\label{supp:implementation}

\subsection{Encoder Information Paths}
\label{supp:encoder}

SAGE integrates spatial-prior, global ViT, and local-view representations to construct a multi-scale feature pyramid for downstream prediction tasks. This section provides additional implementation details of the feature processing, aggregation, and interaction mechanisms.

\textbf{Spatial and contextual modulation.}
For ViT-B, the spatial prior branch extracts patch and CLS tokens from block 9, while the FiLM module aggregates the four register tokens from block 12. All four aggregated ViT feature streams initially share a common stride-16 spatial resolution. Before feature interaction, Laplacian refinement is applied to the first two streams, while register-conditioned FiLM modulation is applied to the fourth stream. The four interaction blocks then sequentially update the SPM query sequence, using the processed ViT features as attention values. In parallel, the original ViT features are preserved through an independent residual pathway, resized to the corresponding output resolutions, and subsequently gated before being added to the updated SPM features. Importantly, the gating mechanism operates on the ViT residual features rather than directly modulating the attention outputs.

\textbf{Local-view fusion.}
Given an input image, a $2\times2$ spatial partition generates four non-overlapping crops without additional resizing before ViT encoding. Each crop is independently processed by the shared ViT backbone. For ViT-B, patch features extracted from blocks 9 and 12 are averaged and reshaped into $32\times32$ feature maps. These maps are then reassembled according to their original spatial positions, forming a $64\times64$ feature map. This crop-based encoding strategy provides complementary contextual views while preserving the original patch sampling stride. The resulting local feature map is subsequently resized to the stride-4 and stride-8 resolutions. At each resolution, the local branch first applies a $3\times3$ convolution followed by normalization and GELU activation. The resulting features are further processed by a dedicated fusion block consisting of a $1\times1$ projection, normalization, and GELU activation. The fusion gate employs $1\times1$ and $3\times3$ convolutions on the concatenated global, local, and absolute-difference features. In the corresponding ablation, removing the local-view pathway also removes its associated fusion operations, while preserving the multi-depth aggregation mechanism.

\paragraph{Multi-depth aggregation.}
For ViT-B, the multi-depth aggregator samples eight transformer blocks, $\{2,3,5,6,8,9,11,12\}$, using one-based backbone indexing. The sampled representations are combined into four weighted feature streams on a shared stride-16 grid, followed by LayerNorm for both spatial and CLS-token features. The aggregation weights are parameterized by learnable logits and normalized across the eight sampled layers using a softmax with temperature $\tau=2.0$. Specifically, let $k\in\{0,\ldots,7\}$ denote the position of a layer within the sampled sequence, rather than its original transformer block index. The initial aggregation logits are defined as
\begin{equation}
W_{s,k}^{\mathrm{init}}=-0.5|k-c_s|,
\end{equation}
where the four aggregation centers are $(c_1,c_2,c_3,c_4)=(1,3,5,7)$. These initialization coordinates correspond to positions in the sampled sequence, not the original backbone block indices. The aggregation weights are learned during training and shared across all input images. Both the sampled block indices and the softmax temperature remain fixed throughout training.

\subsection{Task-specific Interfaces and Input Resolutions}
\label{supp:task_interfaces}

SAGE provides a unified visual encoder for multiple recognition tasks, while retaining task-specific prediction heads to accommodate different output requirements. For COCO object detection and instance segmentation, as well as ADE20K~\cite{zhou2017ade20k} semantic and instance segmentation, we employ MaskDINO as the task-specific decoder. Image classification uses a two-layer MLP head. For human pose estimation, the multi-scale feature pyramid is upsampled to stride 4, followed by a $1\times1$ convolution to predict keypoint heatmaps.  Oriented detection uses an adapted RiO-DETR head.

To ensure fair and consistent comparisons with existing methods, we adopt task-specific input resolutions following the corresponding experimental protocols and comparison settings. Specifically, the input resolutions used in our experiments are summarized in Table~\ref{tab:supp_resolution}.

\begin{table}[t]
\centering
\caption{Input resolutions adopted for different visual recognition tasks.}
\label{tab:supp_resolution}
\small
\begin{tabular}{lcc}
\toprule
Dataset & Task & Input Resolution \\
\midrule
COCO & Object Detection & $1024\times1024$ \\
COCO & Instance Segmentation & $1024\times1024$ \\
COCO & Human Pose Estimation & $192\times256$ \\
ADE20K & Semantic Segmentation & $896\times896$ \\
ADE20K & Instance Segmentation & $1150\times1150$ \\
TinyImageNet & Image Classification & $384\times384$ \\
DIOR-R & Rotated Object Detection & $800\times800$ \\
\bottomrule
\end{tabular}
\end{table}

\begin{table*}[!t]
\centering
\caption{Paired multi-seed evaluation on the fixed 20K-image COCO subset after 17K training updates. All results are obtained using terminal raw checkpoints. The last row reports the mean and sample standard deviation ($n=3$). Gain statistics are calculated from the three within-seed performance differences.}
\label{tab:supp_multiseed}
\setlength{\tabcolsep}{5pt}
\begin{tabular*}{\textwidth}{@{\extracolsep{\fill}}crrrrrr@{}}
\toprule
& \multicolumn{3}{c}{Box AP} & \multicolumn{3}{c}{Mask AP}\\
\cmidrule(lr){2-4}\cmidrule(lr){5-7}
Seed & Baseline & Full SAGE & Gain
     & Baseline & Full SAGE & Gain\\
\midrule
42 & 55.69 & 56.26 & +0.57
   & 48.98 & 49.70 & +0.72\\
43 & 55.79 & 56.46 & +0.67
   & 49.06 & 49.71 & +0.65\\
44 & 55.18 & 56.28 & +1.10
   & 48.68 & 49.65 & +0.96\\
\midrule
Mean $\pm$ SD
   & $55.55\pm0.32$ & $56.33\pm0.11$ & $+0.78\pm0.28$
   & $48.91\pm0.20$ & $49.69\pm0.04$ & $+0.78\pm0.16$\\
\bottomrule
\end{tabular*}
\end{table*}

\subsection{Pre-training Protocol}
\label{supp:pretraining}

\paragraph{Objects365 pre-training.}
Objects365~\cite{shao2019objects365} detector pretraining starts from ViT-B/L weights and uses 365 classes. Both scales use AdamW~\cite{loshchilov2019adamw} (learning rate $10^{-4}$, weight decay 0.05, backbone learning-rate multiplier 0.1). WarmupPolyLR is configured over 1,306,800 iterations for ViT-B and 1,200,000 for ViT-L, with global batches 12 and 16, respectively; the 10-iteration warm-up factor is 1.0. ViT-B uses polynomial power 0.9, \texttt{constant\_ending}=0.01, $1024\times1024$ inputs, large-scale jittering over $[0.1,2.0]$, and an unfrozen ViT backbone. ViT-L uses seed 20260902 and EMA warm-up time constant 2,000.

\paragraph{Downstream initialization.}
Objects365-pretrained ViT-B and ViT-L checkpoints provide the initialization for SAGE-B and SAGE-L, respectively. The task-specific prediction heads and input resolutions are described in Section~\ref{supp:task_interfaces}. In the controlled COCO subset experiments, the Baseline and SAGE variants share the same Objects365-pretrained ViT-B initialization and are trained for 17K updates on the fixed 20K-image subset. ADE20K semantic and instance segmentation initialize ViT-B from raw Objects365 weights and ViT-L from EMA Objects365 weights.

\paragraph{Controlled VFL timing study.}
The detector-pretraining comparison in Section~\ref{supp:vfl} uses a separate 20K-update budget to compare classification-loss schedules under matched initialization and training settings. This short-run comparison is distinct from the pretraining used to produce the initialization checkpoints for the downstream benchmarks. Both schedules are evaluated with raw checkpoints.

\subsection{Benchmark Protocols}
\label{supp:benchmark_protocols}

\paragraph{COCO detection and instance segmentation.}
We train on COCO train2017 and evaluate box and mask AP on val2017 using MaskDINO at $1024\times1024$ input resolution. Full-data training uses 100K iterations, batch sizes of 16--32, AdamW with an initial learning rate of $10^{-4}$ and weight decay 0.05, and WarmupPolyLR. The controlled ablations use a fixed 20K-image training subset for 17K updates with batch size 16 and $1024\times1024$ large-scale jittering. Their terminal raw checkpoints are evaluated on all 5,000 val2017 images; box and mask metrics are obtained from the same checkpoint. The baseline in Table~2 of the main paper (E7) uses ViT-B, an SPM-based one-way adapter, and MaskDINO, without SAGE enhancements and with focal loss. Full-data benchmarks use large-scale jittering over $[0.1,2.0]$, EMA decay 0.9999, and best-EMA evaluation. The cumulative ablation in Table~3 of the main paper also uses EMA evaluation.

\paragraph{ADE20K semantic segmentation.}
The SAGE feature pyramid is decoded by MaskDINO using $896\times896$ inputs. We report mean intersection over union (mIoU) with single-scale inference and no test-time augmentation. Both scales use 150 classes and AdamW (learning rate $10^{-4}$, weight decay 0.05). ViT-B/L use global batches 32/16 and configured WarmupPolyLR horizons of 200K/400K iterations, with 10-iteration warm-up. VFL is disabled; evaluation uses the best EMA checkpoint.

\paragraph{ADE20K instance segmentation.}
MaskDINO predicts instance masks from the SAGE feature pyramid using $1150\times1150$ inputs. Performance is measured by mask AP under single-scale inference without test-time augmentation. Both scales use global batch 16, EMA decay 0.9999, and best-EMA evaluation. ViT-B performs focal-loss training followed by VFL fine-tuning from a focal-stage EMA checkpoint. ViT-L uses a configured 200K-iteration horizon and transitions from focal loss to VFL over 120K--125K.

\paragraph{TinyImageNet classification~\cite{le2015tinyimagenet}.}
We use $384\times384$ inputs and a two-layer MLP classification head. Top-1 and top-5 accuracy are reported for a single model, without ensembling. Both scales use 200 classes, global batch 256, and AdamW with learning rate $10^{-4}$. WarmupCosineLR has a configured 100K-iteration horizon and 1,000-iteration warm-up with factor 1.0. Evaluation uses the best EMA checkpoint.

\paragraph{COCO keypoint estimation.}
We use $192\times256$ inputs. The four feature levels are aligned to stride 4, followed by a $1\times1$ convolution that predicts keypoint heatmaps. Performance is measured by keypoint AP without test-time augmentation. Both scales train for 210 epochs, with learning-rate reductions at epochs 170 and 200. Global batches are 64 for ViT-B and 8 for ViT-L. AdamW uses learning rate $5\times10^{-4}$, weight decay 0.1, layer-wise decay 0.85, and 500-iteration warm-up with factor 0.001. Target-weighted MSE supervises 17 heatmaps of $48\times64$ pixels with Gaussian standard deviation 2. Training uses FP32, gradient clipping at 1.0, and no EMA; evaluation uses the best raw checkpoint. Training augmentations include horizontal flips ($p=0.5$), half-body crops ($p=0.3$), scale jitter $\pm50\%$, rotation $\pm80^{\circ}$, and bounding-box padding factor 1.25.

\paragraph{DIOR-R oriented detection~\cite{cheng2022aopg}.}
We train on the combined training and validation splits and evaluate on the test split using an adapted RiO-DETR head. Inference uses a single $800\times800$ scale without test-time augmentation. AP$_{50}$ and AP$_{75}$ are reported from the same checkpoint selected according to AP$_{50}$. Both scales train for 24K iterations with global batch 32 and AdamW at learning rate $2\times10^{-4}$. The first 1,500 iterations use constant learning rate (warm-up factor 1.0), followed by cosine decay over 22,500 iterations to $10^{-7}$. ViT-B uses weight decay 0.05 and backbone learning-rate multiplier 0.1.

\section{Additional Encoder Ablations and Repeatability}
\label{supp:encoder_evidence}

\subsection{Component Ablations and Saliency Controls}
\label{supp:component_controls}

\begin{figure}[!htbp]
\centering
\includegraphics[width=\columnwidth]{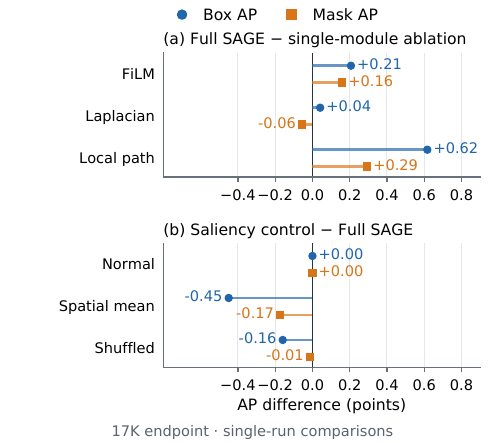}
\caption{Conditional component effects and saliency prior interventions. The upper panel reports the performance difference between the full E1 configuration and each single-component ablation ($\mathrm{E1}-\mathrm{Ablation}$), while the lower panel reports the difference between each saliency control and E1 ($\mathrm{Control}-\mathrm{E1}$). All differences are computed relative to E1, which achieves 56.4594 box AP and 49.7130 mask AP. All results are obtained from single runs without uncertainty estimates.}
\label{fig:supp_component_effects}
\end{figure}

\begin{table}
\centering
\small
\caption{Detector-pretraining trajectories under two VFL training schedules using ViT-B. V1 transitions from focal loss to VFL over iterations 10K--11K.}
\label{tab:supp_startup}
\setlength{\tabcolsep}{3pt}
\begin{tabular*}{\columnwidth}{@{\extracolsep{\fill}}ccccc@{}}
\toprule
Iteration & V0 AP & V1 AP & V0 AP$_{50}$ & V1 AP$_{50}$\\
\midrule
1K  & 0.01  & 0.01  & 0.03  & 0.05\\
2K  & 1.16  & 0.72  & 3.14  & 2.43\\
3K  & 5.68  & 3.87  & 13.42 & 9.82\\
4K  & 10.16 & 8.48  & 23.32 & 20.71\\
5K  & 13.49 & 11.45 & 30.01 & 26.90\\
6K  & 15.62 & 13.48 & 34.13 & 31.06\\
7K  & 17.13 & 15.77 & 37.74 & 35.09\\
8K  & 18.74 & 17.57 & 39.90 & 37.76\\
9K  & 19.14 & 18.92 & 40.58 & 40.16\\
10K & 20.91 & 19.84 & 43.09 & 41.70\\
11K & 21.46 & 21.38 & 44.20 & 44.28\\
12K & 22.13 & 21.93 & 45.04 & 45.12\\
13K & 22.99 & 23.42 & 46.24 & 46.58\\
14K & 23.62 & 23.63 & 46.98 & 46.95\\
15K & 24.13 & 24.11 & 47.14 & 47.43\\
16K & 25.05 & 24.74 & 48.14 & 48.30\\
17K & 25.28 & 25.00 & 48.54 & 48.80\\
18K & 25.35 & 25.24 & 48.74 & 49.03\\
19K & 25.32 & 25.43 & 48.85 & 49.21\\
20K & 25.44 & 25.69 & 48.76 & 49.45\\
\bottomrule
\end{tabular*}
\end{table}

Figure~\ref{fig:supp_component_effects} provides further evidence of the individual contributions of the proposed components and examines the sensitivity of SAGE to different saliency prior configurations. In the upper panel, removing FiLM leads to decreases of 0.21 box AP and 0.16 mask AP relative to the full E1 configuration. Removing the local-view pathway results in larger performance drops of 0.62 box AP and 0.29 mask AP, highlighting its contribution to complementary spatial feature enhancement. In comparison, Laplacian refinement exhibits a relatively small and mixed effect, with a gain of 0.04 box AP but a decrease of 0.06 mask AP when comparing E1 against the corresponding ablation.

\begin{figure*}
\centering
\includegraphics[width=\textwidth]{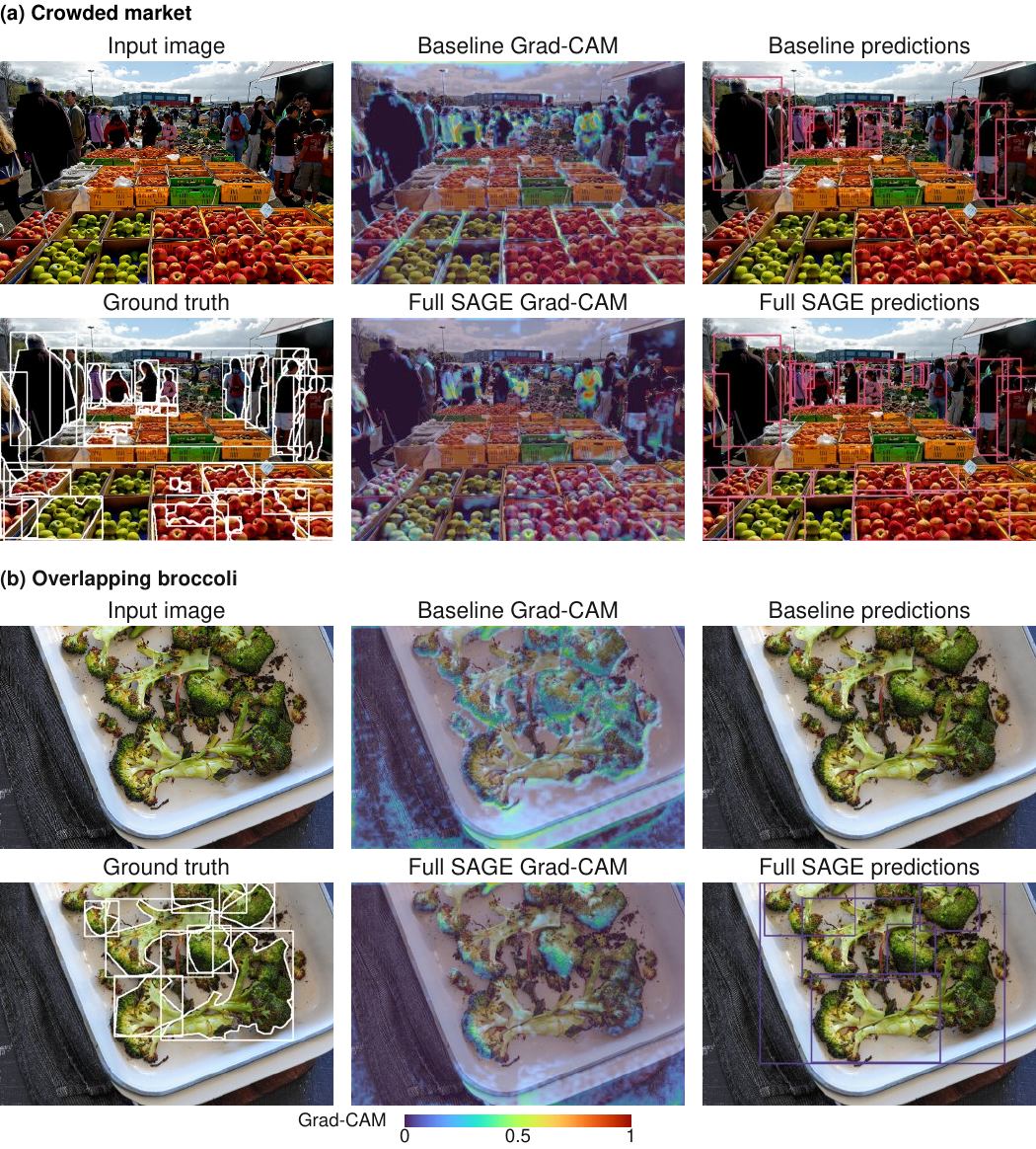}
\caption{Full qualitative comparison on COCO val2017. The market example appears in the main paper; the broccoli example provides an additional view of the prediction and attribution patterns. Colors indicate normalized Grad-CAM values.}
\label{fig:supp_qualitative}
\end{figure*}

We further investigate the role of the proposed saliency prior through two alternative configurations. Specifically, \textit{Normal} denotes the original saliency prior proposed in SAGE, \textit{Spatial mean} replaces the spatially varying saliency prior with its spatial mean, thereby removing its spatial variation, and \textit{Shuffled} randomly permutes the spatial saliency values. As shown in the lower panel, replacing the original prior with its spatial mean leads to decreases of 0.45 box AP and 0.17 mask AP. Randomly shuffling the saliency values also results in performance drops of 0.16 box AP and 0.01 mask AP. These results suggest that the proposed saliency prior provides meaningful benefits beyond spatially uniform or randomly permuted alternatives, supporting the importance of spatially adaptive feature modulation in SAGE.

\subsection{Repeatability Across Training Seeds}
\label{supp:multiseed}

To examine the repeatability of the proposed method, we conduct a paired multi-seed comparison between the Baseline and Full SAGE configurations using three random seeds (42, 43, and 44). The training image subset, backbone initialization, optimizer, and evaluation protocol are kept identical across the corresponding runs. The Baseline employs focal loss throughout training, whereas Full SAGE incorporates the complete proposed architecture and adopts the two-stage training strategy, transitioning from focal loss to VFL over updates 8.5K--9K. Therefore, this comparison evaluates the repeatability of the complete SAGE framework, including both its architectural components and training strategy.

Table~\ref{tab:supp_multiseed} summarizes the results on the fixed 20K-image COCO subset after 17K training updates. All evaluations use the terminal raw checkpoints. For each seed, the performance gain is calculated as the difference between the Full SAGE and Baseline results obtained with the same seed.

Full SAGE consistently outperforms the Baseline across all three seeds on both evaluation metrics. The box AP improvements range from 0.57 to 1.10, while the mask AP improvements range from 0.65 to 0.96, yielding an average gain of approximately 0.78 AP for both metrics. Moreover, Full SAGE exhibits lower observed between-seed standard deviations than the Baseline for both box AP (0.11 vs. 0.32) and mask AP (0.04 vs. 0.20). Although the limited number of seeds does not support a formal statistical significance claim, these paired results provide additional evidence that the performance improvements of SAGE are reproducible across independent training runs.

\subsection{Qualitative Comparison}
\label{supp:qualitative}
The complete qualitative comparison in Figure~\ref{fig:supp_qualitative} includes a crowded market (COCO image 303566) and overlapping broccoli instances (COCO image 527695). For each example, the left column shows the input image above the ground truth; the upper and lower rows show Baseline and Full SAGE Grad-CAM~\cite{selvaraju2017gradcam} maps and predictions, respectively.

In the market scene, SAGE produces more localized Grad-CAM responses over several fruit groups, whereas the Baseline responses extend more broadly along stall and container boundaries. The prediction panels show additional fruit detections with SAGE in this crowded setting. In the broccoli example, the Baseline produces no visible detection in the displayed output, while SAGE detects multiple overlapping instances. SAGE's responses are also more localized within the food region. These examples complement the quantitative encoder ablations by showing differences in feature attribution alongside improved detection coverage under clutter and overlap.

\section{Training Strategy Analyses}
\label{supp:training}

\subsection{Detector-Pretraining VFL Schedules}
\label{supp:vfl}

To examine the effect of different VFL training schedules, we compare two strategies during detector pre-training using the ViT-B backbone. Checkpoints are evaluated every 1K iterations. V0 employs VFL throughout training, whereas V1 uses focal loss for the first 10K iterations, linearly transitions to VFL over iterations 10K--11K, and then uses VFL exclusively. Table~\ref{tab:supp_startup} reports the evaluation trajectories.

V0 achieves higher AP and AP$_{50}$ throughout iterations 2K--10K. Following the transition to VFL, V1 obtains higher AP$_{50}$ at nine of the ten evaluation points between iterations 11K and 20K, while AP alternates between the two strategies. At 20K iterations, V1 outperforms V0 by 0.25 AP and 0.69 AP$_{50}$. These results indicate that, in this detector-pretraining setting, the two-stage VFL schedule achieves better final detection performance than applying VFL from the beginning.

\begin{table}
\centering
\small
\caption{Training controls under the 17K-iteration subset protocol. ``2-stage'' denotes the transition from focal loss to VFL over iterations 8.5K--9K; ``VFL@0'' applies VFL from the beginning of training. Modulation Schedule indicates whether the time-varying auxiliary modulation coefficient is used. Freeze refers to aggregation-logit freezing.}
\label{tab:supp_subset}
\setlength{\tabcolsep}{3pt}
\begin{tabular*}{\columnwidth}{@{\extracolsep{\fill}}llccrr@{}}
\toprule
ID & Loss & Modulation Schedule & Freeze & Box AP & Mask AP\\
\midrule
T1 & 2-stage & $\checkmark$ & $\checkmark$ & \textbf{56.46} & 49.71\\
T2 & 2-stage & $\times$ & $\checkmark$ & 56.36 & 49.67\\
T3 & 2-stage & $\checkmark$ & $\times$ & 56.44 & \textbf{49.78}\\
T4 & VFL@0 & $\checkmark$ & $\checkmark$ & 56.27 & 49.56\\
T5 & VFL@0 & $\times$ & $\checkmark$ & 56.36 & 49.64\\
\bottomrule
\end{tabular*}
\end{table}

\begin{table*}
\centering
\small
\caption{Extended COCO val2017 comparisons. ViT-Adapter-L uses BEiTv2 pretraining. The Co-DETR Swin-L configuration uses IN-22K pretraining and single-scale inference. The ViT-L configuration uses EVA-02 initialization followed by Objects365 training and is evaluated with test-time augmentation. All SAGE results in this and subsequent comparison tables use single-scale inference without TTA. Parameter counts refer to the backbone.}
\label{tab:supp_coco}
\setlength{\tabcolsep}{5pt}
\begin{tabular*}{\textwidth}{@{\extracolsep{\fill}}llrlrr@{}}
\toprule
Method & Backbone & Params & Pretraining / Inference & Box AP & Mask AP\\
\midrule
D-FINE-X~\cite{peng2025dfine} & HGNetv2-B5 & 30M & IN-1K + O365 & 59.3 & --\\
RF-DETR-2XL~\cite{robinson2026rfdetr} & DINOv2-B & 86M & DINOv2 + O365 & 60.1 & --\\
RF-DETR-Seg-2XL~\cite{robinson2026rfdetr} & DINOv2-S & -- & DINOv2 + O365 & -- & 49.9\\
ViTDet~\cite{li2022vitdet} & ViT-H & 632M & IN-1K (MAE) & 58.7 & 50.9\\
MaskDINO (O365)~\cite{li2023maskdino} & Swin-L & 197M & IN-22K + O365 & -- & 54.5\\
ViT-Adapter-L~\cite{chen2023vitadapter} & ViT-L & 304M & BEiTv2 & 58.8 & 51.1\\
Co-DETR (single-scale)~\cite{zong2023codetr} & Swin-L & 197M & IN-22K; no TTA & 60.7 & --\\
Co-DETR (TTA)~\cite{zong2023codetr} & ViT-L & 304M & EVA-02 + O365; multi-scale TTA & 65.9 & --\\
\midrule
SAGE-B & ViT-B & 86M & O365; single-scale & 61.1 & 54.4\\
SAGE-L & ViT-L & 304M & O365; single-scale & \textbf{62.8} & \textbf{55.3}\\
\bottomrule
\end{tabular*}
\end{table*}

\begin{table*}
\centering
\small
\caption{Extended ADE20K semantic segmentation comparisons. All mIoU values use single-scale inference. Parameter counts refer to the backbone. Image--text-derived supervision and additional task-specific data are indicated separately.}
\label{tab:supp_ade20k_semseg}
\setlength{\tabcolsep}{4pt}
\begin{tabular*}{\textwidth}{@{\extracolsep{\fill}}llrlr@{}}
\toprule
Method & Backbone & Params & Pretraining / Extra Data & mIoU\\
\midrule
\multicolumn{5}{l}{\textit{Image-only pretrained references}}\\
Mask2Former~\cite{cheng2022mask2former}
& Swin-L & 197M & IN-22K & 56.1\\
ViT-Adapter + UPerNet~\cite{chen2023vitadapter}
& ViT-L & 304M & IN-22K (BEiT) & 58.0\\
MaskDINO~\cite{li2023maskdino}
& Swin-L & 197M & IN-22K + O365 & 59.5\\
OneFormer~\cite{jain2023oneformer}
& Swin-L & 197M & IN-22K & 58.3\\
ViT-Adapter + Mask2Former~\cite{chen2023vitadapter}
& ViT-L & 304M & IN-22K (BEiT) + COCO-Stuff & 59.4\\
RevCol~\cite{cai2023revcol}
& RevCol-H & 2.1B & 168M images & 60.4\\
DINOv3~\cite{simeoni2025dinov3}
& ViT-7B (frozen) & 7B & LVD-1689M + COCO-Stuff + Hypersim & 62.6\\
\midrule
\multicolumn{5}{l}{\textit{Image--text-pretrained references}}\\
UPerNet~\cite{wang2023internimage}
& InternImage-H & 1.08B & M3I, image--text 427M & 59.9\\
Mask2Former~\cite{wang2023internimage}
& InternImage-H & 1.08B & M3I 427M + COCO-Stuff & 62.5\\
\midrule
\multicolumn{5}{l}{\textit{SAGE: image-only pretraining}}\\
SAGE-B
& ViT-B & 86M & O365 & 58.7\\
SAGE-L
& ViT-L & 304M & O365 & \textbf{60.3}\\
\bottomrule
\end{tabular*}
\end{table*}

\subsection{Auxiliary Modulation Schedule and Aggregation Freezing}
\label{supp:training_controls}

Let $p=t/T$ denote the normalized training progress, where $t$ is the current iteration and $T$ is the total number of training iterations. The auxiliary modulation coefficient is defined as
\begin{equation}
\lambda(p)=
\begin{cases}
\tfrac12[1-\cos(\pi p/r_{\mathrm{up}})],
&0\le p<r_{\mathrm{up}},\\
1,
&r_{\mathrm{up}}\le p<r_{\mathrm{down}},\\
\tfrac12[1+\cos(\pi\frac{p-r_{\mathrm{down}}}{1-r_{\mathrm{down}}})],
&r_{\mathrm{down}}\le p\le1.
\end{cases}
\end{equation}

For the 17K subset experiments, the final modulation decay starts at iteration 10,200, corresponding to $r_{\mathrm{down}}=0.60$. The aggregation logits are frozen at this boundary in the reference configurations. Constant-modulation controls set $\lambda\equiv1$ while retaining the same freezing boundary, whereas the no-freeze control changes only the aggregation-logit freezing strategy. The detector classification-loss schedule is controlled independently.

Table~\ref{tab:supp_subset} summarizes the complementary training controls under the 17K subset protocol. The results show that the two-stage VFL strategy provides consistent benefits under both constant and scheduled auxiliary modulation. Under constant modulation, two-stage VFL yields the same box AP and a 0.03-point gain in mask AP compared with applying VFL from the beginning (T2 vs. T5). With the scheduled auxiliary modulation, the two-stage strategy improves box and mask AP by 0.19 and 0.15 points, respectively (T1 vs. T4). This larger improvement suggests that the scheduled auxiliary modulation provides a more favorable setting for the two-stage VFL strategy.

Removing aggregation-logit freezing changes box and mask AP by $-0.02$ and $+0.07$, respectively (T3 vs. T1), indicating different effects on the two metrics.

\section{Extended Cross-Task Results}
\label{supp:cross_task}

This section provides extended cross-task comparisons with additional methods and configurations beyond those reported in the main paper. We distinguish image-only pretraining from image--text-derived supervision, including supervision transferred from vision--language teachers. We also report relevant differences in backbone scale, pretraining data, additional task-specific training data, and inference protocols to provide context for interpreting the results. Unless otherwise specified, parameter counts refer to the backbone rather than the complete model. SAGE-B and SAGE-L use ViT-B and ViT-L backbones with 86M and 304M parameters, respectively, initialized with Objects365 pretraining.

\subsection{COCO Object Detection and Instance Segmentation}
\label{supp:COCO1}

Table~\ref{tab:supp_coco} extends the COCO comparison with additional backbone and inference configurations. ViT-Adapter-L uses BEiTv2 pretraining. We also report two Co-DETR configurations: the standard Swin-L configuration pretrained on IN-22K, which achieves 60.7 box AP without TTA, and an additional ViT-L configuration initialized with EVA-02 and trained on Objects365, which achieves 65.9 box AP with multi-scale test-time augmentation.

Under the standard single-scale evaluation protocol, SAGE-B and SAGE-L achieve 61.1 and 62.8 box AP, respectively, both outperforming the reported Co-DETR result of 60.7. The higher Co-DETR score of 65.9 is obtained with an augmented inference protocol involving multi-scale TTA, which incurs additional test-time computation and is not directly comparable to the single-scale, no-TTA evaluation used by SAGE. Moreover, the two Co-DETR results employ different backbones and pretraining configurations, so their performance difference should not be attributed solely to TTA.

These comparisons highlight the importance of considering inference protocols alongside the reported accuracy. Under the single-scale setting adopted in the main paper, SAGE-L achieves higher detection AP than the reported standard Co-DETR configuration, while the TTA-enhanced Co-DETR result provides an additional reference under a more computationally intensive inference setting. SAGE also achieves 54.4 and 55.3 mask AP with SAGE-B and SAGE-L, respectively.

\begin{table*}
\centering
\small
\caption{Extended ADE20K instance segmentation comparisons. ``+COCO'' denotes additional COCO training. Parameter counts refer to the backbone configuration.}
\label{tab:supp_ade20k_inst}
\setlength{\tabcolsep}{3pt}
\begin{tabular*}{\textwidth}{@{\extracolsep{\fill}}llrr@{}}
\toprule
Method / Extra Training & Backbone & Params & Mask AP\\
\midrule
Mask2Former + DiNAT~\cite{jain2023oneformer} & DiNAT-L & 200M & 35.4\\
OneFormer~\cite{jain2023oneformer} & DiNAT-L & 200M & 37.1\\
OneFormer + ViT-P + COCO~\cite{jain2023oneformer,shahabodini2026vitp} & DiNAT-L + ViT-P & 300M & 40.7\\
MP-Former~\cite{zhang2023mpformer} & Swin-L & 197M & 35.3\\
OneFormer~\cite{jain2023oneformer} & Swin-L & 197M & 37.8\\
OneFormer + ViT-P~\cite{shahabodini2026vitp} & DiNAT-L + ViT-P & 300M & 37.8\\
OneFormer + COCO~\cite{jain2023oneformer} & DiNAT-L & 200M & 40.2\\
\midrule
SAGE-B & ViT-B & 86M & 43.3\\
SAGE-L & ViT-L & 304M & \textbf{44.9}\\
\bottomrule
\end{tabular*}
\end{table*}

\begin{table*}
\centering
\caption{Additional DIOR-R references and SAGE results. AP$_{75}$ corresponds to the checkpoint selected by AP$_{50}$. Parameter counts refer to the backbone. RingMoE-KC uses RingMOSS, RT-DETRv2 uses IN-1K, and SAGE uses Objects365 pretraining.}
\label{tab:supp_diorr}
\setlength{\tabcolsep}{6pt}
\begin{tabular*}{\textwidth}{@{\extracolsep{\fill}}llrrr@{}}
\toprule
Method & Backbone & Params & AP$_{50}$ & AP$_{75}$\\
\midrule
Oriented R-CNN~\cite{zhang2025skysensev2} & SkySense V2 MoE & 1994M & 75.29 & --\\
RT-DETRv2-OBB-L~\cite{lv2024rtdetrv2} & HGNetV2-B4 & 13.6M & 75.69 & --\\
Oriented R-CNN~\cite{li2025ringmoe} & RingMoE-KC & 1000M & 76.04 & --\\
YOLO26x-OBB~\cite{wang2026yolo26,hu2026riodetr} & YOLO26x & 56M & 76.48 & --\\
RT-DETRv2-OBB-X~\cite{lv2024rtdetrv2} & HGNetV2-B5 & 33M & 76.52 & --\\
RiO-DETR-X~\cite{hu2026riodetr} & HGNetV2-B5 & 33M & 77.43 & --\\
\midrule
SAGE-B & ViT-B & 86M & 78.44 & 51.96\\
SAGE-L & ViT-L & 304M & \textbf{81.10} & \textbf{56.86}\\
\bottomrule
\end{tabular*}
\end{table*}

\begin{table*}
\centering
\small
\caption{Extended TinyImageNet classification comparison. ViT-L/16, CaiT-S/36, and Swin-L/4 use IN-22K pretraining; DeiT and OCD use IN-1K; CLAdapter uses IN-21K; SAGE uses Objects365 pretraining. OCD ensemble denotes a five-model ensemble.}
\label{tab:supp_tinyimagenet}
\setlength{\tabcolsep}{3pt}
\begin{tabular*}{\textwidth}{@{\extracolsep{\fill}}llrrr@{}}
\toprule
Method & Backbone & Params & Top-1 (\%) & Top-5 (\%)\\
\midrule
ViT-L/16~\cite{dosovitskiy2021vit,huynh2022vit} & ViT-L/16 & 304M & 86.43 & 97.30\\
CaiT-S/36~\cite{touvron2021cait,huynh2022vit} & CaiT-S/36 & $\approx68$M & 86.74 & 97.42\\
DeiT-B/16-D~\cite{touvron2021deit,huynh2022vit} & DeiT-B/16-D & 87M & 87.29 & 96.80\\
OCD (DeiT-B, single)~\cite{lutati2023ocd} & DeiT-B/16-D & $\approx87$M & 90.80 & --\\
Swin-L/4~\cite{liu2021swin,huynh2022vit} & Swin-L/4 & 196M & 91.35 & 98.50\\
OCD (DeiT-B, ensemble $\times5$) & DeiT-B/16-D & $\approx87$M$\times5$ & 92.00 & --\\
CLAdapter~\cite{li2025cladapter} & ViT-L & 304M & 94.21 & --\\
\midrule
SAGE-B & ViT-B & 86M & 91.46 & 98.49\\
SAGE-L & ViT-L & 304M & \textbf{94.76} & \textbf{99.11}\\
\bottomrule
\end{tabular*}
\end{table*}

\subsection{ADE20K Semantic Segmentation}
\label{supp:ADE20K2}

Table~\ref{tab:supp_ade20k_semseg} extends the ADE20K semantic segmentation comparison with methods using different pretraining scales and supervision sources. Among the image-only references, RevCol-H achieves 60.4 single-scale mIoU with a 2.1B-parameter backbone pretrained on 168M images. DINOv3 reports 62.6 mIoU using a frozen 7B backbone together with a trained adapter and decoder, with additional task-specific data including COCO-Stuff and Hypersim. These configurations operate at substantially larger model and data scales than SAGE.

Image--text-derived supervision provides another source of strong representations. InternImage-H, with a 1.08B-parameter backbone and M3I pretraining on 427M image--text pairs, obtains 59.9 mIoU with UPerNet and 62.5 mIoU with Mask2Former and additional COCO-Stuff data.

In comparison, SAGE-B achieves 58.7 mIoU with an 86M-parameter ViT-B backbone, while SAGE-L obtains 60.3 mIoU with a 304M-parameter ViT-L backbone. These results provide competitive performance with substantially smaller backbones than several large-scale references, while using Objects365 image pretraining. The comparisons should be interpreted in the context of their different pretraining sources, model scales, and additional training data.

\begin{table}
\centering
\caption{Extended COCO keypoint estimation comparison. Parameter counts refer to the backbone. SAGE uses Objects365 initialization, while the ViTPose and RTMPose references use IN-1K pretraining.}
\label{tab:supp_keypoint}
\setlength{\tabcolsep}{4pt}
\begin{tabular*}{\columnwidth}{@{\extracolsep{\fill}}lrrl@{}}
\toprule
Method & Backbone & Params & AP\\
\midrule
RTMPose-L~\cite{jiang2023rtmpose}
& CSPNeXt-L & 26M & 75.8\\
ViTPose-B~\cite{xu2022vitpose}
& ViT-B & 86M & 75.8\\
ViTPose-L~\cite{xu2022vitpose}
& ViT-L & 306M & 78.3\\
ViTPose-H~\cite{xu2022vitpose}
& ViT-H & 631M & \textbf{79.1}\\
\midrule
SAGE-B
& ViT-B & 86M & 76.0\\
SAGE-L
& ViT-L & 304M & \underline{78.5}\\
\bottomrule
\end{tabular*}
\end{table}

\subsection{ADE20K Instance Segmentation}
\label{supp:ADE20K3}

Table~\ref{tab:supp_ade20k_inst} provides additional ADE20K instance segmentation comparisons. The listed reference methods include configurations with different backbones and additional COCO training. In particular, OneFormer with ViT-P uses additional COCO data, providing a stronger reference than configurations trained without this extra data.

SAGE-B and SAGE-L achieve 43.3 and 44.9 mask AP, respectively. Both results are higher than the additional reference configurations listed in the table, including OneFormer with ViT-P and extra COCO training. This comparison highlights the effectiveness of the SAGE encoder for instance-level recognition without requiring the additional COCO training used by that reference configuration.

\subsection{DIOR-R Oriented Detection}
\label{supp:DIOR-R}

Table~\ref{tab:supp_diorr} extends the DIOR-R comparison with additional oriented detection methods and reports both AP$_{50}$ and AP$_{75}$ for SAGE. Evaluation is conducted on the DIOR-R test set after training on train+val, using single-scale $800\times800$ inference without TTA. The AP$_{75}$ values correspond to the checkpoints selected according to AP$_{50}$.

SAGE-B achieves 78.44 AP$_{50}$ and 51.96 AP$_{75}$, while SAGE-L obtains 81.10 AP$_{50}$ and 56.86 AP$_{75}$. The additional references also cover substantially different model scales and architectures. SkySense V2 MoE and RingMoE-KC use backbones with approximately 1.994B and 1B parameters, respectively, considerably larger than the SAGE backbones. RT-DETRv2-OBB-L uses HGNetV2-B4 with 13.6M parameters and IN-1K pretraining. These results provide additional reference points across different architectures, parameter scales, and pretraining settings.

\subsection{TinyImageNet Classification}
\label{supp:TinyImageNet}

Table~\ref{tab:supp_tinyimagenet} extends the TinyImageNet comparison by reporting both top-1 and top-5 accuracy, together with single-model and ensemble configurations. SAGE-B and SAGE-L achieve 91.46\% and 94.76\% top-1 accuracy, respectively, with corresponding top-5 accuracies of 98.49\% and 99.11\%.

Among the additional references, OCD with a five-model ensemble reaches 92.00\% top-1 accuracy. This result uses an ensemble of five models and is therefore not directly comparable to the single-model SAGE results. The table also includes methods pretrained on IN-22K or IN-21K, whereas SAGE uses Objects365 pretraining. These differences in pretraining data and inference configuration should be considered when interpreting the results.

\subsection{COCO Keypoint Estimation}
\label{supp:COCO_keypoint}

Table~\ref{tab:supp_keypoint} extends the COCO keypoint estimation comparison with additional model scales. ViTPose provides a useful reference across ViT-B, ViT-L, and ViT-H backbones, while RTMPose-L represents a compact CNN-based alternative. SAGE-B achieves 76.0 AP, exceeding the reported RTMPose-L and ViTPose-B results of 75.8 AP. SAGE-L obtains 78.5 AP, outperforming ViTPose-L at 78.3 AP, while remaining below the larger ViTPose-H configuration at 79.1 AP.

Notably, ViTPose-H uses a 631M-parameter backbone, more than twice the size of the 304M ViT-L backbone in SAGE-L. These comparisons demonstrate that SAGE achieves competitive keypoint estimation performance across both backbone scales, with SAGE-L surpassing the ViT-L-based reference while using a smaller backbone than ViTPose-H.

\end{document}